\pdfoutput=1

\documentclass[11pt]{article}

\usepackage{acl}

\usepackage{times}
\usepackage{latexsym}

\usepackage[T1]{fontenc}

\usepackage[utf8]{inputenc}

\usepackage{microtype}

\usepackage{inconsolata}

\usepackage{graphicx}

\usepackage{url}
\usepackage{array}
\usepackage{colortbl}  
\usepackage{threeparttable}
\usepackage{multirow}
\usepackage{here}
\usepackage{MnSymbol}

\usepackage{pifont} 
\newcommand{\cmark}{\ding{51}} 
\newcommand{\xmark}{\ding{55}} 

\title{Can AI Examine Novelty of Patents?: Novelty Evaluation Based on the Correspondence between Patent Claim and Prior Art}

\author{Hayato IKOMA \and Teruko MITAMURA \\
  Language Technology Institute, SCS \\
  Carnegie Mellon University \\
  \texttt{ikomahayato@jcom.zaq.ne.jp} \\}

\begin{document}
\maketitle
\begin{abstract}
Assessing the novelty of patent claims is a critical yet challenging task traditionally performed by patent examiners. While advancements in NLP have enabled progress in various patent-related tasks, novelty assessment remains unexplored. This paper introduces a novel challenge by evaluating the ability of large language models (LLMs) to assess patent novelty by comparing claims with cited prior art documents, following the process similar to that of patent examiners done. We present the first dataset specifically designed for novelty evaluation, derived from real patent examination cases, and analyze the capabilities of LLMs to address this task. Our study reveals that while classification models struggle to effectively assess novelty, generative models make predictions with a reasonable level of accuracy, and their explanations are accurate enough to understand the relationship between the target patent and prior art. These findings demonstrate the potential of LLMs to assist in patent evaluation, reducing the workload for both examiners and applicants. Our contributions highlight the limitations of current models and provide a foundation for improving AI-driven patent analysis through advanced models and refined datasets.
\end{abstract}

\section{Introduction} \label{sec:introduction}
Evaluating the patentability and assessing the novelty of an invention are among the most challenging and critical tasks in patent analysis. Traditionally, these tasks have been carried out exclusively by patent examiners, who undergo extensive training and possess deep, specialized knowledge. Meanwhile, patents have been explored from various perspectives in the field of NLP research. This has led to the proposal of numerous tasks, including support tasks, classification, and retrieval tasks \cite{krestel2021survey}. However, novelty assessment remains a particularly challenging task, leaving ample room for valuable research in this area.

\begin{figure}[t]
  \includegraphics[width=\columnwidth]{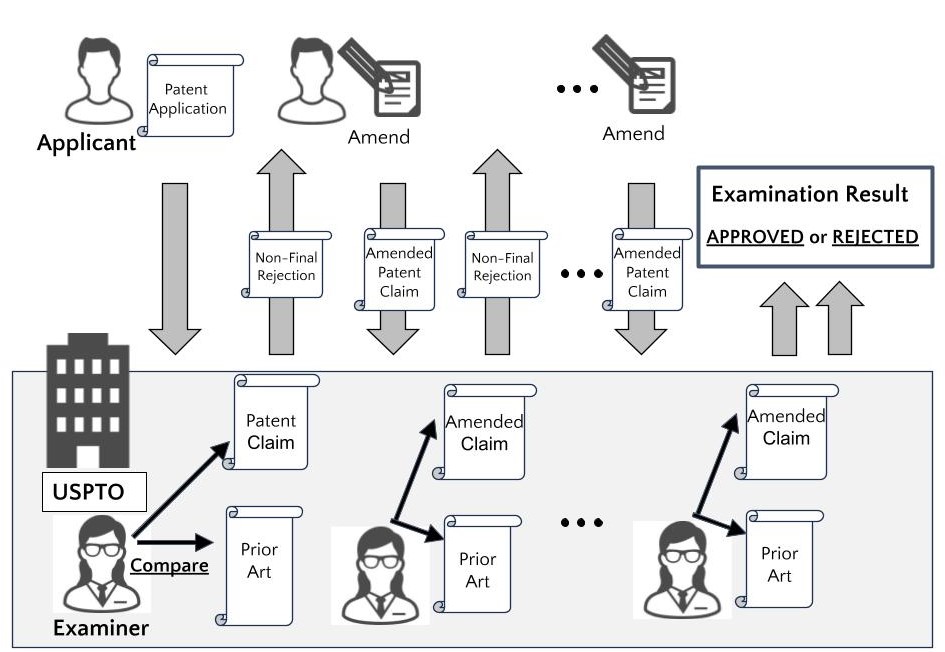}
  \caption{An Overview of Patent Examination Process: Examiners compare patent claims with prior art and issue a non-final rejection if grounds for rejection are found. Applicants can amend the claims, and the examiner compare them with prior art again. This process repeats until a decision is made to approve or reject the patent.}
  \label{fig:experiments}
\end{figure}

One of the most important and difficult points in the task of evaluating novelty is comparing the invention with prior arts. Figure  \ref{fig:experiments} illustrates the actual examination process. The examiner of the USPTO(United States Patent and Trademark Office) compares the invention to the prior arts to judge whether  it satisfies the novelty requirements under patent law\footnote{In the United States, the following law code defines novelty;\\35 U.S. Code § 102 - Conditions for patentability; novelty}. If examiner finds a single prior art document, which contains all elements of the claimed invention in the submit patent document, the invention is regarded as non-novel. In this case, before making final decision, they send an documents which cites the most similar prior art document and represents the result of comparison and the reason of rejection. This document is called "Non-Final Rejection". The applicants have the opportunity to amend the claim and dispute the rejection. If the every elements of the original or amended patent claim are not found in the cited prior art documents, the invention is regarded as novel and approved to be patented. If the applicants finally failed to amend the patent claim to satisfy the novelty requirements, the patents are rejected. This process demands substantial technological and legal expertise from both applicants (or their attorneys) and examiners, consuming significant time and resources. If this process can be replaced by AI, or if AI can provide advice or suggestions, it would significantly reduce the workload and assist individuals without expertise in performing tasks related to obtaining patents.

The main purpose of this research is evaluating the ability of the latest AI to assess the presence or absence of novelty based on an analysis similar to the patent examination method, involving specific claim texts and cited documents. In recent years, large language models (LLMs) have rapidly developed, and their application to tasks in the patent and legal domains has been extensively explored \cite{lee2019patentbert, koreeda2021contractnli}. Recent models that handle longer input tokens \cite{beltagy2020longformer, MetaLlama3} have published, which enables us to challenge this theme. Datasets for novelty evaluation, consisting of pairs of patent claims and descriptions from cited documents based on actual patent examination results, has not been created before. This study is the first attempt to evaluate the novelty assessment capabilities of LLMs, which have advanced reasoning abilities for long-text inputs, using a dataset based on actual examination results. Clarifying these capabilities enables us to more appropriately employing LLMs in the patent acquisition and evaluation processes, potentially reducing the effort and costs for companies and patent offices. 

Our contributions and findings are as follows:
\begin{itemize}
    \item We propose a task and dataset for novelty assessment based on actual patent cases and their examination documents. 
    \item The classification models used in this study show limited ability to assess novelty. Further analysis is demanded for more accurate evaluation. 
    \item The generation models demonstrate a certain capability to assess novelty, along with the validity and characteristics of their explanations.
\end{itemize}

\section{Task}
\begin{figure*}[t]
\centering
  \includegraphics[width=13cm]{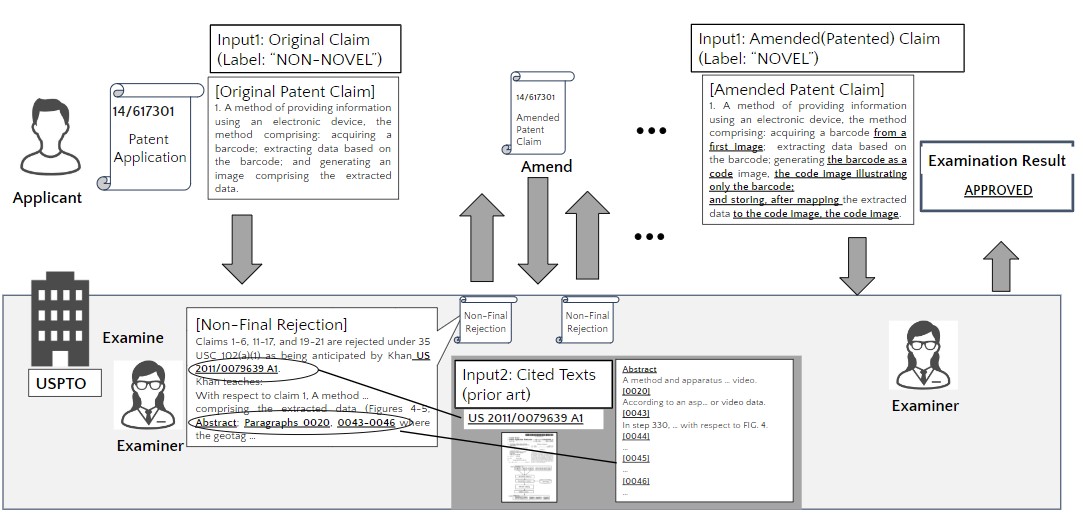}
  \caption{Entire patent examination process and patent documents from which input data is extracted: the Input Claim is came from Original Claim(label "Non-Novel") or Amended Claim(label: "Novel"). the Input Cited Texts is came from cited prior art documents. "14/617301" and "US 2011/0079639 A1" are example document numbers; Application Number and Publication Number.}
  \label{fig:entirepatexam}
\end{figure*}

To evaluate the ability of LLM models to assess novelty, it is essential to align the tasks with actual patent practices. Therefore, we propose a new binary classification task, which lead models to determine presence or absence of novelty in the claim from the relationship between patent claims and cited texts of prior art in the actual Non-Final Rejection. We name this two input condition "Claim-Cited Texts (C-T) input".

The innovative aspect of our proposed task is the following: it evaluates novelty from the relationship between the claim text and the cited texts. This is fundamentally different from previous research which assessed patent values based on attributes such as the applicant or technical field, or that which predicted patentability through syntactic and semantic analysis of claims. Our research reflects the actual examination process, focusing on the interaction between these two inputs.

Another novel feature is the use of Non-Final Rejection documents to directly extract relevant parts from the whole documents. This allows the model to accurately identify which parts of sentences correspond to elements of the claim. In previous studies, Non-Final Rejection documents have not been utilized, but they provide strong evidence for the presence or absence of novelty. Numerous records of Non-final rejection documents have been stored in the USPTO database. This makes it easy to scale up the dataset for our proposed task in future research.

For defining the task, the source of input claim and cited texts of prior arts is a notable issue.As Figure \ref{fig:experiments} shows, in the actual examination process, the patent claims and the cited references are compared many times under the applicant's multiple amendments, and the final decision is whether the patent is granted or rejected. In many cases, the amended claims have eliminated the ground of rejection of novelty, and other grounds of rejection, such as non-obviousness are often at issue. Therefore, patent claims that are held to lack of novelty are often the original claims before the amendment. On the other hand, in the case of granting a patent, specific descriptions of references similar to the patent claims are not explicitly provided (only document names, such as patent publication numbers, are provided). Therefore, we changed the source of the patent claim between the approved.

Figure \ref{fig:entirepatexam} shows the source of input claim and input cited texts. The task targets patent applications that have been notified of a reason for rejection due to lack of novelty. The data for the "Non-Novel" label consists of two inputs: (1) The text of the original claim that has been notified of Non-Final rejection due to lack of novelty, and (2) the texts of cited paragraphs in the referenced patent publication document stated in the first Non-Final rejection document. The data for the "Novel" label consists of two inputs;(1)The text of the final amended claim that were modified by the applicant after the non-final rejection and considered to be novel, and (2) the texts of cited paragraphs (same as the data for the "Non-Novel" label).

In addition to C-T input, we also introduced single input condition using only claim text(named "Only Claim(C) input"). It has been found that single passage models could achieve a good result with some multi-document/passage task, and it is recommended to test with single input condition for precise evaluation \cite{poliak-etal-2018-hypothesis,kaushik-lipton-2018-much}. Our proposed task is assessed novelty from whether patent claim is patented or not,which is a kind of multi document/passage task. In addition to it, some researches(e.g. \citealp{suzgun2024harvard, gao2022towards}) reported that patentability can be predicted in a certain accuracy from only claim texts. Therefore, we conducted test with C input condition for more adequate evaluation.

\section{Datasets}

\subsection{Data Source}
We synthesized multiple data sources from the USPTO APIs and Google patent public data \cite{googleGPPD}. To improve compatibility with HUPD \cite{suzgun2024harvard}, every patent document collected is also contained in HUPD. The data covers the period from 2014 to 2015 (applied), and the technical field is limited to IPC \textbf{G06F}\footnote{ELECTRIC DIGITAL DATA PROCESSING}, which had the largest occupation among all IPC subclasses in 2014\footnote{The occupation is investigated by OECD stat \cite{OECD_stat}}.

\subsection{Pre-processing}
Most patent applications have multiple claims. The input claim for "Non-Novel" data is the first claim that was notified in the Non-final rejection as lacking novelty. The input claim for the "Novel" data is the first claim that has patentability\footnote{There are two types of claims: independent claims and dependent claims, but only independent claims were used.}.

The cited text data used as input was obtained from the extracted content of the complete document. In the Non-Final rejection process, prior patent documents are cited to demonstrate the existence of prior art, and the most relevant paragraphs are indicated. The cited text data was identified by paragraph numbers as indicated in the rejection documents using regular expressions (see Figure \ref{fig:entirepatexam}, specifically "Input2 Cited Texts").

The data were divided into 80\% for training, 10\% for evaluation and 10\% for testing. However, analysis of the collected data revealed an undesirable bias: longer claims tend to be labeled "Novel". Because this bias might result in improper learning during model training, we removed it from the training and evaluation data in the following way: 1) dividing the set of data into groups based on the number of words in the claims (e.g., 100-110 words), and 2) each group was then randomly filtered, keeping the label ratio at 1:1. The statistics of the dataset before and after pre-processing are shown in Table \ref{tab:dataset_stats}.

\section{Methodology}

Our task represents one of the newest and most innovative challenges in multi-passage classification. We have employed cutting-edge models and methods specifically developed for this type of classification.

We approached our task in two ways: 1) using a classification head on the top linear layer;2) using text generation with several types of prompts. Although the first approach is common for encoder-only models, it has been reported that the decoder-only models, such as Llama2 perform effectively in classification tasks using a classification head at the top(e.g., \citet{li2023label})\footnote{The classification head for Llama models uses the last token to perform the classification \cite{hgllama2}}.

For classifying with text generation, we designed prompts that mirror real examination processes. Learning paradigms includes zero-shot, few-shot(2-shot), and supervised fine-tuning.

\renewcommand{\arraystretch}{0.8}
\begin{table}
\centering
\tabcolsep 4pt
\begin{tabular}{llll}
\hline
\multicolumn{4}{c}{\textbf{Amount of Patents}} \\
\hline
\multicolumn{2}{l}{Total Patents} & \multicolumn{2}{l}{3975} \\
\multicolumn{2}{l}{Train:Valid:Test} & \multicolumn{2}{l}{3180:397:398} \\
\multicolumn{2}{l}{(Pre-Processed)} & \multicolumn{2}{l}{1396:166:398} \\
\hline
\multicolumn{4}{c}{\textbf{Word Counts}} \\
\hline
& \multicolumn{1}{c}{Label} & \multicolumn{2}{c}{Ave.(Std.)} \\
\hline
\multicolumn{4}{c}{\textbf{Claim}(Train)} \\
& Novel & \multicolumn{2}{l}{216(104.2)} \\
& Non-Novel & \multicolumn{2}{l}{112(55.5)} \\
\multicolumn{4}{c}{(Pre-Processed)} \\
& Novel & \multicolumn{2}{l}{151.5(56.8)} \\
& Non-Novel & \multicolumn{2}{l}{151.3(56.6)} \\
\hline
\multicolumn{4}{c}{\textbf{Cited Text}} \\
& Novel & \multicolumn{2}{l}{917(988.8)} \\
& Non-Novel & \multicolumn{2}{l}{903(872.6)} \\
\hline
\multicolumn{4}{c}{\textbf{Label Ratio}} \\
\hline
& Novel:Non-Novel & \multicolumn{2}{l}{1:1} \\
\hline
\end{tabular}
\caption{Dataset statistics. Pre-processing was applied to balance the dataset and remove length bias in claims for training and validation sets.}
\label{tab:dataset_stats}
\end{table}
\renewcommand{\arraystretch}{1}

\subsection{Models}
Because the proposed task is long document classification task(longer than 512 words), We have to choose the models which are able to handle longer input tokens. The chosen models are: Longformer \cite{beltagy2020longformer}, Llama2\cite{touvron2023llama}, Llama3 (Meta(2024)\cite{MetaLlama3}), GPT-4o\cite{openAIGPT4o} to assess them with their architecture and parameter size. 

\noindent\textbf{Longformer}: Longformer is a Encoder-only model pretrained for MLM on long documents\cite{beltagy2020longformer}. We chose \textit{longformer-base-4096}, which is based on the RoBERTa checkpoint and has a maximum sequence length of 4,096 tokens.

\noindent\textbf{Llama2, Llama3}: As the representative open-source decoder-only model with long context window, we chose Llama2, Llama3. The context window of Llama2 is 4096, that of Llama3 is 8192, and they were pretrained with adequately long documents\cite{touvron2023llama}. We used Llama2 7B, 13B and Llama3 8B, 70B. For fine-tuning of Llama models, we used QLoRA PEFT methods\cite{dettmers2024qlora}. Due to computer resource constraints, we only conducted zero-shot and few-shot tests on the 70B model. When we test model with the classification head, pretrained versions were used. For classifying with text generation, instruction-tuned versions were used.

\noindent\textbf{GPT-4o}: As one of the latest and the largest size model, we selected GPT-4o. Same as Llama-3 70B, we only conducted zero-shot and few-shot tests.

\subsection{Prompting}
The Prompt for the task guide models to classify labels from the novelty of the claims. Specifically, prompt for Claim-Cited texts inputs instructs models to classify labels by determining whether every and each claim elements are found in the cited texts, either expressly or inherently. This instruction is extracted and revised from the § 2131 of the Manual of PATENT EXAMINING PROCEDURE\cite{mpepe131}.

We designed distinct prompts based on specific experimental settings, including the presence or absence of explanation of the output reason (Predict-only or Explain-Predict). These prompts are adapted to the type of learning paradigms(zero-shot, few-shot, and Supervised Fine-Tuning(SFT)).

The Predict-Only prompt is designed to classify a label based on the input and prohibits providing any explanation about the classification process.  This condition was introduced to make comparison with the Predict-Explanation prompt case. Current text generation models tend to automatically generate explanations even when instructed to simply "Predict labels". Therefore, under this condition, we explicitly prohibited providing any explanation, as in "Predict labels without any explanation."

The Explain-Predict prompt orders to generate explanation for the analysis of claims and cited texts before generating the classified label. This prompt condition is introduced not only to verify whether such explanation leads to improve the performance of classification\footnote{Although it is well-known that making models to explain the reason for their prediction attributes their ability(e.g.\cite{kojima2022large},\cite{lampinen2022can}), some researches show that explanation does not always improve performances in certain cases(e.g.\cite{huang2023can}).}, but also to ascertain whether the model is effectively classifying based on the correspondence between claims and the cited texts. For the few-shot paradigm, we created sample explanation based on the actual patent claims and its Non-Final Rejection documents. For Only Claim input, we made sample explanation from the original patent claim(label:"Non-Novel"), or the amended claim(label:"Novel"). For Claim-Cited Texts inputs, we used Non-Final Rejection documents in addition to the patent claims. 

In the few-shot paradigm, the prompt incorporated two exemplars: one labeled 'Novel' and the other labeled 'Non-Novel', which were fame from the training data. For the SFT paradigm, Predict-Only prompt was employed, due to the impracticality of generating explanations for each instance in the training dataset. 

\section{Result}

In the experiment, we began by conducting Human Testing to establish a benchmark for the task accuracy proposed in this study, as well as to identify specific challenges involved. Following this, we presented the results of the task across different models and experimental conditions. For the most successful experimental conditions, we performed a detailed Qualitative Analysis to examine the nature of the outputs. This approach enabled us to evaluate the extent to which current high-parameter models can predict novelty and to highlight their limitations.

\subsection{Human Testing}
To verify the nature of the task, one of the authors categorized examples under two input conditions. 20 examples were extracted randomly from the test data, in which the undesirable correlation between word count and labels was eliminated to ensure task validity. With Only Claim input, human accuracy was \textbf{0.650} (13/20), and with claim-cited texts input, the accuracy was \textbf{0.600} (12/20). These results illustrate the difficulty of this task.

From the context of claim texts, it is possible to make predictions with a certain level of accuracy based on the technical content and the level of abstraction. On the other hand, predicting based on the relationship between the claims and the cited texts requires accurately assessing whether all elements of the claims are described in the cited texts. In actual examinations, examiners can grasp the details of claim elements and cited text descriptions from various sources, including drawings and detailed technical explanations. Nevertheless, in this task, the input text is limited, and errors can arise from misunderstandings of the claims and cited texts or from a lack of specialized knowledge in the technical field. Moreover, differences in judgment between the classifier and actual examiners may occur, such as in determining whether a claim falls under the broader concept of a cited text description. 
Therefore, when analyzing the results of this human testing experiment, it is important to note that the input related to the content of prior art does not necessarily lead to an improvement in accuracy. Despite this, it is crucial to recognize that these results are based on a very limited sample size (n = 1) and a small number of examples (20), which may not fully represent broader trends or outcomes. 

\subsection{Longformer, Llama2, 3(with the classification head)}

\begin{table}[t]
\centering
\small
\tabcolsep 2pt
\begin{threeparttable}
\begin{tabular}{lcccc}
\hline
  \multicolumn{1}{c}{(Average Acc.(5 times))}    & \multicolumn{1}{c}{Longformer}&\multicolumn{2}{c}{Llama2\tnote{a}}&\multicolumn{1}{c}{Llama3\tnote{a}}\\
\   Input   &    & 7B & 13B & 8B \\
\hline
Claim(C) & 0.563 & 0.650 & \underline{0.658} & 0.590 \\
Claim-Cited Texts(C-T) & 0.503 & 0.507 & 0.480 & \underline{0.515} \\
\hline
\end{tabular}
\begin{tablenotes}
\item[a]Llama2,3 in the table are instruction-tuned versions.
\end{tablenotes}
\end{threeparttable}
\caption{Results of Longformer, Llama2,3 with the classification head. In this case, Llama2,3  are instruction-tuned versions.}
\label{tab:result1}
\end{table}

The performance of models with the classification head is presented in Table\ref{tab:result1}. Llama models with the classification head consistently outperform Longformer models across most input conditions. This observation aligns with the experiment results that Llama2 models with the classification head outperform encoder-only models, such as BERT, RoBERTa, in classification tasks. Considering Only Claim input, the top-performing model achieved an accuracy exceeding 0.65. However, A substantial decline in accuracy was observed for the claim-cited text comparison condition, where performance approached 0.5, which means randomly predicted. These results suggest that even considering the difficulty of the task, the classification model is not effectively predicting labels based on the relationship between the claim and the cited text. This raises the possibility that the cited text input is actually functioning as noise.

\subsection{Llama2,3, GPT-4o}\label{sec:generation}
\begin{table}[t]
\centering
\small
\tabcolsep 3pt
\begin{threeparttable}
\begin{tabular}{lllccccc}
\hline
\multicolumn{3}{l}{(Acc.(1 time))}    &\multicolumn{1}{c}{Llama2\tnote{a}}&\multicolumn{2}{c}{Llama3\tnote{a}}&\multicolumn{1}{c}{GPT-4o}\\
\ Input&&  &  13B & 8B & 70B &  & \\
\hline
\multirow{5}{*}{C}&\multirow{3}{*}{P-O}&0-shot  & 0.638 & 0.510 & 0.550 & \underline{0.653} \\
&&few-shot  & 0.593 & 0.525 & \underline{0.701} & 0.691 \\
&&SFT  & \underline{0.651} & 0.593 & - & - \\
\cline{2-7}
&\multirow{2}{*}{E-P}&0-shot  & 0.595 & \underline{0.661} & 0.530 & 0.656 \\
&&few-shot  & 0.595 & 0.621 & 0.704 & \underline{0.729} \\
\arrayrulecolor{black}\hline  
\multirow{5}{*}{C-T}&\multirow{3}{*}{P-O}&0-shot & 0.536 & 0.553 & \underline{0.581} & 0.548 \\
&&few-shot  & 0.548 & 0.509 & \underline{0.593} & 0.563 \\
&&SFT & 0.395 & \underline{0.488} & - & - \\
\cline{2-7}
&\multirow{2}{*}{E-P}&0-shot & 0.516 & 0.490 & 0.558 & \underline{0.580} \\
&&few-shot  & 0.425 & 0.510 & \underline{0.624} & 0.538 \\
\arrayrulecolor{black}\hline  
\end{tabular}
\end{threeparttable}
\caption{Results of Llama2,3 and GPT-4o. In this case, Llama2,3 are instruction-tuned versions.}
\label{tab:result2}
\end{table}

Table\ref{tab:result2} shows the performance of generation models. Considering Only Claim input, although there are some exceptions, improvement of accuracy is observed by opting few-shot or SFT rather than 0-shot. In particular, Llama3 70B and GPT-4o achieved accuracy exceeding human predictions both Predict-Only and Explain-Predict prompts. This results suggest that improvements in prompts, using larger parameter models, and fine-tuning lead models can more accurately capture the characteristics of claim texts, thereby enhancing the prediction of patentability.

The results of claim-cited texts input suggest that models with smaller parameter, such as Llama2 13B and Llama3 8B, appear to predict insufficiently. Considering the SFT results, the results of llama2 13B are significantly low, because it often outputs both "Novel" and "Non-Novel" for one instance. In this case, we invalidated the labeling and did not count it towards either label. The output of Llama3 8B with SFT is comparably stable, but the accuracy is less than 0.5. These results suggest that they has not been successfully trained. Most of other results of Llama2 13B and Llama3 8B are close to 0.5. However, Llama3 70B, a model with large-parameter shows relatively high performance among all models, particularly showing performance surpassing human benchmarks with Explanation-Predict and few-shot condition.

\begin{table}[t]
\centering
\small
\tabcolsep 1pt
\begin{threeparttable}
\begin{tabular}{lllcccccc}
\hline
&& Label\tnote{a}&n(True)\tnote{b}&N(Pred) &Pre.&Rec.&F1&ACC.\\
\hline
\multirow{4}{*}{P-O}&\multirow{2}{*}{0-shot} & Nov &197&303& 0.55 & 0.85& 0.67 &0.85\\
&& Non-N &199 & 93&0.68 & 0.32 & 0.43&0.32 \\
\cline{2-9}
&\multirow{2}{*}{few-shot} & Nov &196&229& 0.58 & 0.68& 0.63& 0.68\\
&& Non-N &197&163& 0.61 & 0.51 & 0.56 &0.51\\
\hline
\multirow{4}{*}{E-P}&\multirow{2}{*}{0-shot} & Nov &197&370& 0.53 & 0.99& 0.69&0.99 \\
& &Non-N &199&26& \textbf{0.96} & 0.13 & 0.22 &0.13\\
\cline{2-9}
&\multirow{2}{*}{few-shot} & Nov &195&311&0.58& 0.92& 0.71 &0.92\\
&& Non-N &196&79& \textbf{0.81}& 0.33 & 0.47 &0.33\\
\arrayrulecolor{black}\hline  
\end{tabular}
\begin{tablenotes} 
\item[a] The "Nov/Non-N" labels mean "Novel/Non-Novel".
\item[b] Data loss from context window restrictions slightly alters the sum of n(True).
\end{tablenotes}
\end{threeparttable}
\caption{Detailed Results of Llama3 70B(Input:\textbf{C-T} )}
\label{tab:llama3_70B_detail}
\end{table}

Table\ref{tab:llama3_70B_detail} shows a bias towards "Novel" predictions for llama3 70B using Claim-Cited Texts input. Although this trend is observed across other generative models, no bias was seen or bias is not consistent in the results with Claim-Only input or those of Classification Models. Additionally, high precision for "Non-Novel" labels was noted in Explain-Predict prompts. This heightened precision for "Non-Novel" label was also evident with GPT-4o. These findings suggest several conclusions regarding the capabilities of large-parameter models in this task. Firstly, the unique bias indicates that  generative models assess relationship of Claim-Cited texts inputs rather than relying solely on claim texts. Secondly, considering the bias, models excel at discerning differences between inputs but have difficulties in assessing content similarity. Finally, the excellent precision for the "Non-Novel" label of large-parameter models suggests that the model has ability making accurate predictions for instances where the correspondence is clear.

\subsection{Qualitative Analysis}

\begin{table*}[t]
\centering
\small
\tabcolsep 3pt
\begin{threeparttable}
\begin{tabular}{cccccl}
\hline
No.&Patent Application Num. & Label(True) & Label(Pred) & Result & Note \\
\hline
1&14 / 636567 & Non-Novel & Novel & \xmark & \underline{Missed the content implied in the text} \\
2&14 / 719240 & Novel & Novel & \cmark & Reasonable explanation \\
3&14 / 640115 & Novel & Novel & \cmark & Reasonable explanation \\
4&14 / 707621 & Non-Novel & Non-Novel & \cmark & Reasonable explanation \\
5&14 / 704111 & Novel & Novel & \cmark & Reasonable explanation \\
6&14 / 769475 & Novel & Novel & \cmark & Reasonable explanation \\
7&14 / 432202 & Non-Novel & Novel & \xmark & \underline{Missed understanding the correspondence}\\
8&14 / 733718 & Novel & Novel & \cmark & Reasonable explanation \\
9&14 / 743459 & Novel & Novel & \cmark & Reasonable explanation \\
10&14 / 922522 & Novel & Novel & \cmark & Reasonable explanation \\
11&14 / 729102 & Non-Novel & Novel & \xmark & \underline{Failed to grasp the full content of cited texts} \\
12&14 / 834266 & Non-Novel & Non-Novel & \cmark & Reasonable explanation \\
13&14 / 929526 & Non-Novel & Novel & \xmark & \underline{Missed the content implied in the text} \\
14&14 / 830073 & Novel & Novel & \cmark & \underline{Missed understanding the correspondence} \\
15&14 / 790199 & Novel & Novel & \cmark & Reasonable explanation \\
16&14 / 799261 & Novel & Novel & \cmark & Reasonable explanation \\
17&14 / 684839 & Novel & Novel & \cmark & Reasonable explanation \\
18&14 / 716077 & Novel & Novel & \cmark & Reasonable explanation \\
19&14 / 607537 & Non-Novel & Novel & \xmark & \underline{Missed the content implied in the text} \\
20&14 / 770005 & Non-Novel & Novel & \xmark & \underline{Failed to grasp the full content of cited texts} \\
\hline
\end{tabular}
\end{threeparttable}
\caption{Qualitative Analysis for Outputs of Llama3 70B:Input condition is Claim-Cited texts, Prompt Condition is Expain-Predict, Learning Paradim is few-shot.}
\label{tab:quali_alalnysis}
\end{table*}

Based on the detailed evaluation of results, We conducted a qualitative analysis of model explanations, focusing on the highest-performing configuration (Llama3 70B; Explain-Predict prompt;Few-shot). We extracted 20 test data instances and the corresponding model outputs randomly. Table \ref{tab:quali_alalnysis} provides an overview of the analysis. Most of the explanations, especially those for correct labels, are reasonable and explain each element of the claim as well as the corresponding or differing parts of the cited texts. No hallucinations, defined as descriptions based on neither the claim nor the cited texts, were found. For "Non-Novel" label instances, most explanations are similar to the actual Non-Final Rejection documents. For "Novel" label instances, the explanations highlight the novel parts of the claim that are not found in the cited texts, most of which are additional parts in the amended claim.

Explanations for failed labels are not entirely failures. While some explanations could not grasp the full content of the cited texts, others missed parts of the correspondence relationships. It is particularly challenging for models to understand complex relationships, such as those involving hypernyms and hyponyms and/or those grounded in specialized technical knowledge. For example, in the No.14 instance, the patent claim represents an invention of \textit{"a method for executing instructions of a transaction in a transaction execution (TX) computing environment"} which includes an element of \textit{"\underline{a latent modification instruction (LMI)} within the transaction"}. The cited texts describe speculative instructions using a prefix called a "speculative instruction prefix". Because this prefix instruction is an example of LMI, examiner assert the correspondence between them in the actual examining process. However, the model failed to recognize this hypernyms and hyponyms relationship between them in this instance.

Additionally, when the cited texts imply relevant details rather than stating them explicitly, models struggle to confirm correspondences or similarities. In the No.1 instance, the patent claim represents an invention of \textit{"a memory system"} which includes an element of \textit{"a second control unit configured to ... \underline{create parity} from the information loaded in the second memory, \underline{store the created parity}"}. In the cited texts, things related to parity are not described explicitly, but it is refereed to transfer data of the RAM to the data register. Although this description implies to create and store parity in the data register, model missed it. However, it is noteworthy that some explanations in other instances do successfully assert the claim elements that are implied in the cited texts. 

This analysis aligns with the insights from Section~\ref{sec:generation} and demonstrates that a large-parameter generation model can explain patent validity at a high level through comparisons between patent claims and prior art. This suggests that the model's ability to generate explanations provides valuable insights into the patent validation process, even with its current limitations.

\section{Related Work}
 In the research field of patentablity evaluation, ~\citet{suzgun2024harvard} developed a large-scale patent dataset (HUPD) and one of the proposed task using HUPD is evaluating the ability of predicting patentability from claim text or abstracts. ~\citet{gao2022towards} suggested a method for predicting patent approvals using information about novelty and proposed a framework, AISeer, that unifies the document classifier with handcrafted features, particularly time-dependent novelty scores. ~\citet{risch2020patentmatch} published a training dataset containing pairs of claims and corresponding text passages from cited patent documents. Although this dataset is similar to ours, their text passages were divided into paragraphs rather than providing the entire content. Related to our approach to compare each element of claims to cited texts, ~\citet{schmitt2023assessment} suggested a mathematical-logical approach for assessing patentability by comparing the feature combinations of patent claims with the pertinent prior art. In aspects of comparison between patent claims and patent descriptions, ~\citet{hashimoto2023hunt} proposed a new task to make comparisons between the claim and its description paragraphs to extract unclaimed embodiments. However, this is the first challenge to evaluating abilities of models, including generative models, using whole cited texts which are longer than 1k amount of words.

\section{Conclusion and Future Work}
This research aimed to evaluate LLM models, focusing on their understanding of the relationship between patent claim contents and prior art. We have revealed characteristics of both classification and generative models.

We found that the traditional architecture of classification models, which sets a classification head on a linear layer, did not effectively capture the corresponding relationship between claims and cited texts in our dataset. Our task and dataset are based on the patentability of each patent document, which is strongly influenced by the syntactic and semantic characteristics of the claims. To focus on evaluating correspondence, it is considerable to create a new task and dataset featuring fragmented claim elements annotated with their correspondences based on the actual examination process.

Our another finding was that large-parameter generation models, such as Llama 3 70B, have the ability to predict labels with certain accuracy, especially under few-shot conditions. Moreover, with limited instructions based on the actual examination manual for human examiners, the model generated beneficial explanations in many cases. Additionally, we clarified the limitations of the current model and trends in the explanations, which enhance the utility of generation models in the evaluation of patents and prior art.

However, because our exploration of prompts is still limited, there is room to improve instructions and examples. Changing prompts might lead models to alter their output trends, and the approach of exploring prompts may potentially improve the accuracy and explainability of the model. We used 2 examples for few-shot conditions due to the limited context window of Llama 2. For future research, using only models with large context windows, such as Llama 3 or GPT-4, increasing the number of examples, and adjusting the proportion of labels in examples seems to be a better approach.

\section*{Acknowledgement}
This research was sponsored by the Japan Patent Office. We would like to thank Kimihiro Hasegawa, Adithya Pratapa, Zhi-Qi Cheng and Sang Choe for their valuable advice and contributions to this study.

\bibliography{Hayato_CMU.bib}

\appendix

\setcounter{table}{0}
\renewcommand{\thetable}{A\arabic{table}}
\setcounter{figure}{0}
\renewcommand{\thefigure}{A\arabic{figure}}
\section{Data Collection and Preparation Procedures}
\label{sec:app_Data_collection}
We collected data from the USPTO APIs and the Google Patent Public Dataset. All patent applications included in the dataset must have received a Non-final Rejection for the "lack of novelty" reason. Original claim texts for the "Non-Novel" label are derived from published patent applications, while claim texts for the "Novel" label come from granted patent applications. Notably, applications for the "Non-Novel" label are published before receiving a Non-Final Rejection. Considering these conditions, we obtained the data using the following methods:
\begin{enumerate}
    \item \textbf{Creating Lists}: We created lists of patent applications that received notifications of lack of novelty and were classified under the \textbf{G06F} IPC subclass, using the Patent Examination Research Dataset and Patent Application Office actions data.
    \item \textbf{Eliminating Undesirable Patents}: We eliminated undesirable patents from the prosecution history retrieved via the Patent Examination Data System API. The elimination criteria were: (i) patents published after the first Non-Final Rejection notification; (ii) patents amended between publication and the first Non-Final Rejection notification.
    \item \textbf{Retrieving Non-Final Rejection Texts}: We retrieved the texts of Non-Final Rejections through the USPTO Office Action Text Retrieval API.
    \item \textbf{Identifying Key Information}: We identified key information, such as the number of rejected claims, the publication numbers of referenced patent documents, and the cited paragraph numbers, from the rejection texts using regular expressions. Information on cited figures was discarded.
    \item \textbf{Retrieving Texts of Claims and Referenced Documents}: We retrieved the texts of targeted patent claims and referenced documents using the Google Patent Public Dataset. This dataset was chosen because it offers both claims and descriptions in a structured form, enabling extraction based on claim numbers or paragraph numbers. We retrieved claims from either pre-grant publications or Granted publications depending on the label.
    \item \textbf{Extracting Claim Texts and Cited Texts}: We extracted claim texts and cited texts from the documents retrieved in step (5). The extracted claim has the lowest claim number among the claims in each patent application.
\end{enumerate}

\section{Hyperparameters}
\begin{table}[t]
\centering
\tabcolsep 2pt
\begin{threeparttable}
\begin{tabular}{lcc}
\hline
Models&lr&Batch  \\
\hline
Longformer & 2e-5 & 8   \\
Llama2(with CL Head\tnote{a} ) & 2e-5 & 4  \\
Llama3(with CL Head\tnote{a} ) & 2e-5 & 4  \\
Llama2(SFT\tnote{b} ) & 2e-5 & 4  \\
Llama3(SFT\tnote{b} ) & 2e-5 & 4  \\
\hline

\end{tabular}
\begin{tablenotes}
\item[a]with the classification head
\item[b]Supervised Fine-Tuning
\end{tablenotes}
\end{threeparttable}
\caption{Hyper parameters}
\label{tab:hyperparameters}
\end{table}

\begin{table}[t]
\centering
\tabcolsep 1pt
\begin{threeparttable}
\begin{tabular}{lccccc}
\hline
Models&Max Length&Temp.&Top-k&Top-p&RP\tnote{a}  \\
\hline
Llama2 7B & 512 & 1 &50&1&1 \\
Llama2 13B & 512 & 1&50&1& 1 \\
Llama3 8B & 1024 &  1&50&1&1 \\
Llama3 70B & 1024 &1&50&1& 1 \\
GPT-4o &-&1&-\tnote{b}&1&1\\
\hline

\end{tabular}
\begin{tablenotes}
\item[a]Reputation Penalty
\item[b]The Top-k parameter is not adjustable for GPT-4o.
\end{tablenotes}
\end{threeparttable}
\caption{Hyper Parameters for Text Generation}
\label{tab:hp_text_gene}
\end{table}

The hyperparameters used for fine-tuning are shown in Table \ref{tab:hyperparameters}. We used Early Stopping with "eval\_loss" as the metric in the fine-tuning process. As a result, in most trials, the final epoch ranged between 2 and 4.\\
The hyperparameters for text generation are shown in Table \ref{tab:hp_text_gene}. The parameters except for "Max Length" are the default values of the Transformers library from Hugging Face.

\section{Prompt}

\begin{figure*}[htbp]
\begin{tabular}{cc}
\begin{minipage}{.2\textwidth}
\centering
\includegraphics[width=1\linewidth]{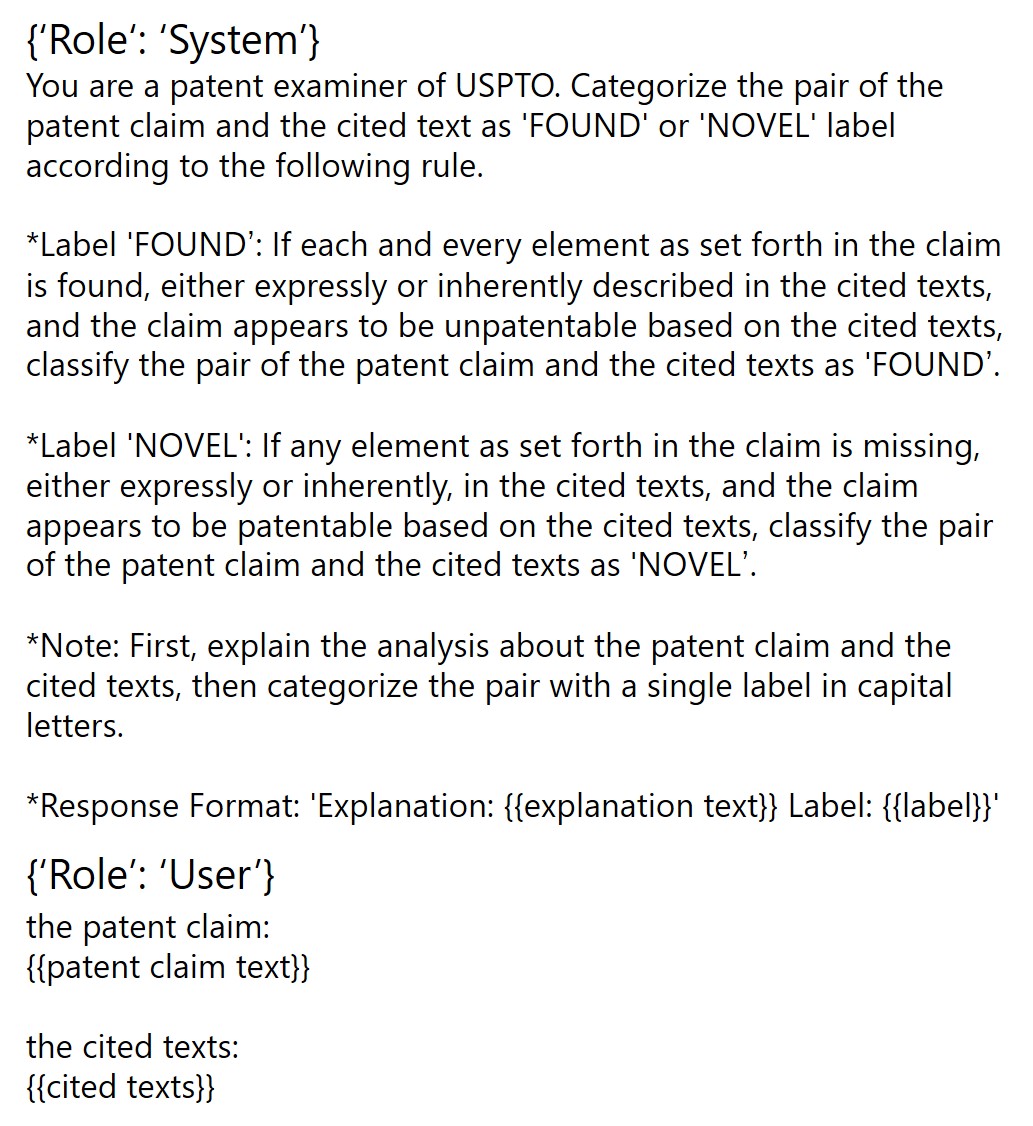}
\caption{Prompt for the E-P condition and few-shot paradigm. In the experiments, claim text data and cited texts data are inserted in the User role prompts.}
\label{fig:prompts}
\end{minipage}
\hspace{0.06\textwidth}
\begin{minipage}{.57\textwidth}
\vspace{3mm}\centering
\includegraphics[width=1\linewidth]{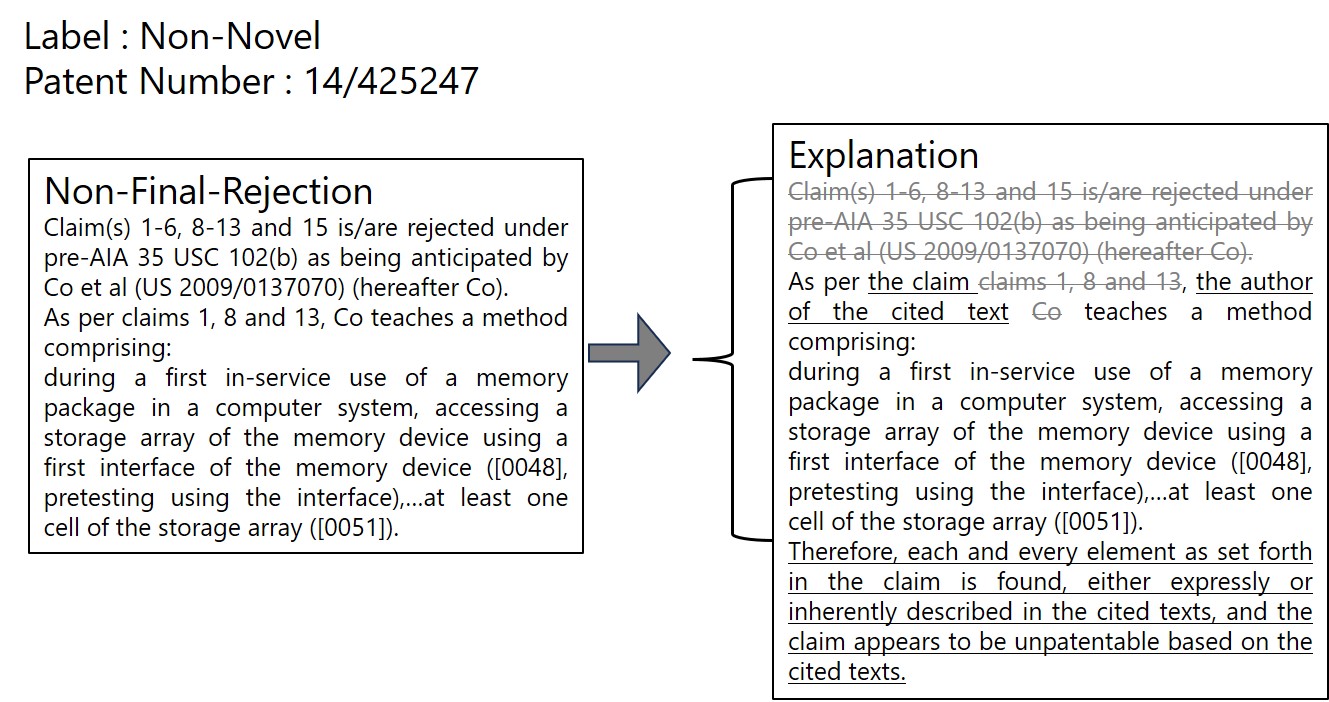}
\caption{Explanation created from the Non-Final Rejection in the prompt for the "Non-Novel" label.}
\label{fig:exp_non_novel}
\end{minipage}
\end{tabular}
\end{figure*}

\begin{figure*}[t]
  \centering
  \includegraphics[width=0.8\linewidth]{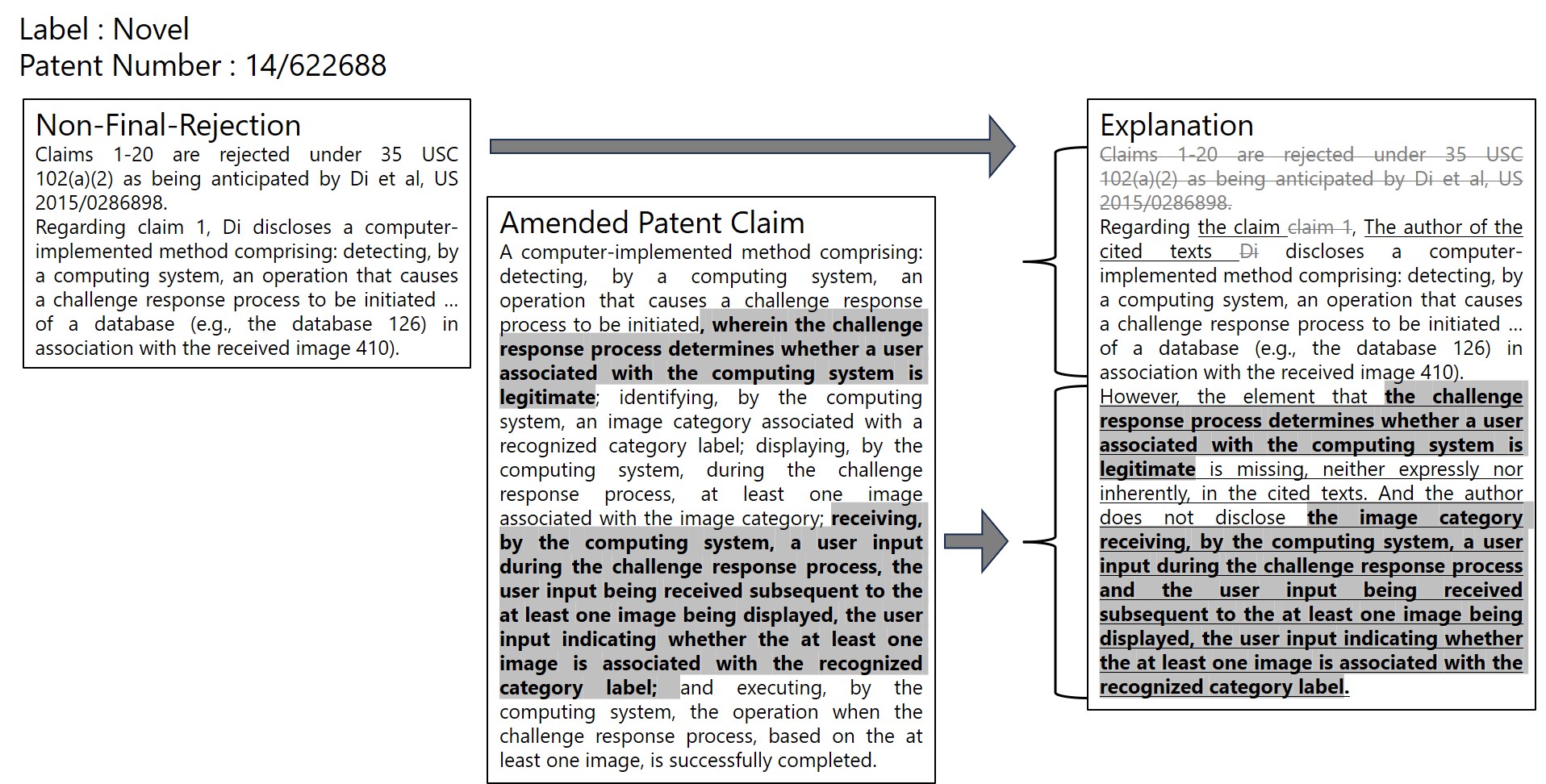} 
  \caption {Explanation created from the Non-Final Rejection in the prompt for the "Non-Novel" label. The additional parts of the amended claim are highlighted in gray.}
\label{fig:exp_novel}
\end{figure*}

The prompt for the E-P condition and few-shot paradigm is shown in Fig\ref{fig:prompts}. The labeling rules in the introduction are based on the § 2131 of the Manual of PATENT EXAMINING PROCEDURE, \textit{“A claim is anticipated only if each and every element as set forth in the claim is found, either expressly or inherently described, in a single prior art reference.” Verdegaal Bros. v. Union Oil Co. of California, 814 F.2d 628, 631, 2 USPQ2d 1051, 1053 (Fed. Cir. 1987)."}.As illustrated in the Figure \ref{fig:prompts}, this sentence was revised to be appropriate as an instruction for this task. \\
When we used the original label names "Novel/Non-Novel" for the prompts, the models were confused and frequently classified the incorrect label with an appropriate explanation. Therefore, we used the label names "Novel/Found" for the prompts. \\
The explanation examples were created from the Non-Final Rejection and amended claims, as shown in Figure \ref{fig:exp_non_novel}, \ref{fig:exp_novel}. Unnecessary information, such as inventor names and  publication numbers, was removed or anonymized in the texts of the Non-Final Rejection. For the "Non-Novel" label data, the sentence \textit{"Therefore, each and every element as set forth in the claim is found, either expressly or inherently described in the cited texts, and the claim appears to be unpatentable based on the cited texts."} was added at the end. For the "Novel" label data, we added some sentences to explain the novel points of the amended claims based on the additional parts in the claims.

\end{document}